\documentclass{svproc}

\usepackage{url}

\usepackage[T1]{fontenc}
\usepackage{graphicx}
\usepackage{algorithm}
\usepackage{algorithmicx}
\usepackage{caption}
\usepackage{subcaption}
\usepackage[noend]{algpseudocode}
\usepackage{amsmath}
\usepackage{sidecap}

\begin{document}
\mainmatter

\title{Securing Federated Learning in Robot Swarms using Blockchain Technology}

\titlerunning{Blockchain Federated Learning in Robot Swarms}

\author{Alexandre Pacheco\inst{1}
\and
Sébastien De Vos\inst{1}
\and
Andreagiovanni Reina\inst{1,2,3}
\and
Marco Dorigo\inst{1}
\and
Volker Strobel\inst{1}
}

\authorrunning{Alexandre Pacheco et al.}
\institute{IRIDIA, Universit\'e Libre de Bruxelles, 1050 Brussels, Belgium\\\email{alexandre.melo.pacheco@ulb.be, mdorigo@ulb.ac.be, volker.strobel@ulb.be} \and
CASCB, University of Konstanz, 78464 Konstanz, Germany\\\email{andreagiovanni.reina@uni-konstanz.de}\and Department of Collective Behaviour, Max Planck Institute of Animal Behavior, 78464 Konstanz, Germany
}

\tocauthor{Alexandre Pacheco, Sébastien De Vos, Andreagiovanni Reina, Marco Dorigo, Volker Strobel}

\maketitle

\begin{abstract}
Federated learning is a new approach to distributed machine learning that offers potential advantages such as reducing communication requirements and distributing the costs of training algorithms. Therefore, it could hold great promise in swarm robotics applications. However, federated learning usually requires a centralized server for the aggregation of the models. In this paper, we present a proof-of-concept implementation of federated learning in a robot swarm that does not compromise decentralization. To do so, we use blockchain technology to enable our robot swarm to securely synchronize a shared model that is the aggregation of the individual models without relying on a central server.
We then show that introducing a single malfunctioning robot can, however, heavily disrupt the training process. To prevent such situations, we devise protection mechanisms that are implemented through secure and tamper-proof blockchain smart contracts. Our experiments are conducted in ARGoS, a physics-based simulator for swarm robotics, using the Ethereum blockchain protocol which is executed by each simulated robot. 
\keywords{swarm robotics, blockchain technology, federated learning}
\end{abstract}

\section{Introduction}

In federated learning, each agent performs local training (for example, on a smartphone) on the data it has collected, and shares the trained model with a centralized server that aggregates the individual models. It is a recent approach to distributed machine learning~\cite{McMMooRam-etal:2017} that has the advantage of keeping data local and private, and distributing the costs of training models among the nodes in a distributed network.

Here, we explore the decentralized implementation of federated learning for the purpose of using it in a robot swarm. 
In a swarm robotics implementation of federated learning, each robot of the swarm locally trains a model with the data it collects. However, instead of a central server, a decentralized and secure data structure is required to aggregate the individual models. Virtual stigmergy has been used for this purpose in previous research~\cite{MajSriPin2021:icra,StOVarSvoBel2020:frontiers}; however, this shared data structure is not secure against the wide range of faults and attacks that may occur in real-world deployments of peer-to-peer systems. As an alternative, we employ the Ethereum blockchain~\cite{But2014:techreport} as a decentralized data structure maintained by the robot swarm, as done in previous research~\cite{StrCasDor2018:aamas,StrCasDor2020:frontiers,PacStrDor2020:ants,PacStrReiDor2022:ants,ZhaPacStr-etal2023:iros,StrPacDor2023:sciencerobotics}. Blockchain protocols such as Ethereum have seen widespread adoption on the internet, and their security properties are well documented.

Even though a blockchain protocol can enable the secure synchronization of a database and execution of smart contracts (programs which are executed synchronously by the distributed network), there is yet the problem of securing the federated learning algorithm from the inputs of malfunctioning or malicious robots (Byzantine robots~\cite{StrCasDor2018:aamas}) that may send incorrect individual models. Therefore, using a blockchain serves a dual purpose: as a distributed computation platform that enables the synchronization of an aggregated model; and as a means to implement security mechanisms that address the risks posed by Byzantine robots for model training.

Previous research introduced blockchain-based federated learning in the context of medical data to ensure data privacy and security~\cite{WarSchSha2021:nature}. In our approach, we investigate the specific challenges associated with networks of mobile robots, which are characterized by local communication capabilities and time-critical information processing. Specifically, our robots are connected in a peer-to-peer manner, with each robot functioning as a blockchain node that maintains the blockchain network. Due to the limited communication range of the robots, the network topology changes rapidly, causing information to reach different robots at different times. This work provides a new perspective on how decentralized machine learning can be implemented securely in robot swarms.

We begin by providing an overview of the background and related work in Section~\ref{sec:background}. Next, in Section~\ref{sec:methods}, we present the experimental implementation, including the simulation architecture that we implemented, the environment, the robots and their behaviors, as well as the federated learning framework, the behaviors of Byzantine robots, and the protective measures that we designed. Subsequently, we present the results of each experiment in Section~\ref{sec:results}. Finally, we conclude with a discussion of our study's implications and potential future developments in Section~\ref{sec:discussion}.

\section{Background}
\label{sec:background}

In the Flow-FL system~\cite{MajSriPin2021:icra}, it has been shown that federated learning can be used by a robot swarm to perform decentralized and collective learning of a model for trajectory prediction. Flow-FL uses virtual stigmergy~\cite{PinLeeBel2016:bict}, a shared database system that was tailored for swarm robotics and therefore is capable of managing the network partitioning and packet losses that may occur in these systems. However, virtual stigmergy and Flow-FL do not provide a way to manage conflicts caused by Byzantine robots (e.g., if a robot sends different models to different peers).
In a real-world deployment of federated learning in robot swarms, we can expect conflicts to occur due to unforeseen circumstances, but they may also occur if a robot falls under the control of a malicious agent that attempts to generate conflicts that lead to state divergence across the network. 

We believe that a first step for implementing federated learning in swarm robotics is to ensure synchronization of a conflict-free distributed data structure that accommodates the shared machine learning model. Otherwise the distributed data structure could potentially become a single point-of-failure for the system. 

The second step is to secure the federated learning algorithm from non-conflicting, yet incorrect inputs. Our results (Section \ref{sec:results}) show that Flow-FL~\cite{MajSriPin2021:icra} is vulnerable to the introduction of a single Byzantine robot that sends random model weights, which could occur, for example, as the result of sensor faults, or when a malicious agent gains control of a robot and manipulates the shared model, an attack known as \textit{model poisoning}~\cite{FunYooBes2020:usenix}. Without security mechanisms, federated learning exposes the entire robot swarm to a high risk of failure as soon as a single robot fails.

However, securing federated learning is a significant challenge, since the individual models are learned from private data and it is difficult to evaluate the quality of a model without accessing this data~\cite{xu_verifynet_2020}. As such, aggregation algorithms such as FedAvg~\cite{McMMooRam-etal:2017} may additionally employ security measures such as Sybil protection~\cite{FunYooBes2020:usenix}. Blanchard
et al.~\cite{blanchard_grad_descent_2017} propose a method called Krum that ensures Byzantine fault-tolerance up to a certain number of attackers. Multiple variants have been proposed such as multi-Krum and median-Krum~\cite{de_rango_median_krum}. 

However, these techniques have been implemented through central servers.
To achieve a decentralized implementation, we integrate the federated learning framework with an existing blockchain technology framework~\cite{PacStrDor2020:techreport-001,StrPacDor2023:sciencerobotics}. This integration is chosen because the blockchain mechanism has proven to be effective when dealing with Byzantine robots in swarm robotics~\cite{StrCasDor2020:frontiers,PacStrDor2020:ants,ZhaPacStr-etal2023:iros,DorPacReiStr2024naturereviewstech,Cas2023:frontiersblockchain}. We then show a smart contract implementation that prevents the harm caused by Byzantine robots, enabling the secure federated learning of the shared model.

\section{Methods}
\label{sec:methods}

We use the same simulation environment and robot controllers as the Flow-FL system~\cite{MajSriPin2021:icra}. The robots move in the environment and have access to GPS coordinates. Each robot records trajectory data about other robots it meets. The trajectory data is then used for training a machine learning model: a deep neural network with an LSTM (long short-term memory) layer and a dense layer (we employed the same neural network configuration as in the original Flow-FL paper~\cite{MajSriPin2021:icra}). These individually learned models are then exchanged with local peers through blockchain transactions, and aggregated into a shared model using a smart contract that is maintained via a blockchain that is independently executed by each robot.

\subsection{Simulation Setup}

We use the physics-based simulator ARGoS~\cite{PinTriOGr-etal2012:si}, the Ethereum blockchain software~\cite{But2014:techreport} and the virtualization software Docker~\cite{Mer2014:linux}. The controller of each robot in ARGoS communicates with a corresponding Docker container that hosts an Ethereum node. In this way, the robot controllers can send transactions and interact with the smart contracts, while the synchronization and maintenance of the blockchain are executed in the Docker containers in the background.
Similar setups have been used in previous research~\cite{StrCasDor2018:aamas,StrCasDor2020:frontiers,PacStrDor2020:ants,PacStrReiDor2022:ants,ZhaPacStr-etal2023:iros,StrPacDor2023:sciencerobotics}. The source code for our setup and all experiments is available online~\cite{PacDeVRei-etal2024:zenodo}.

\subsection{Environment}

\begin{SCfigure}[][t]
\centering
    \includegraphics[width=0.41\textwidth, trim={6cm 0.5cm 6cm 0.5cm}, clip]{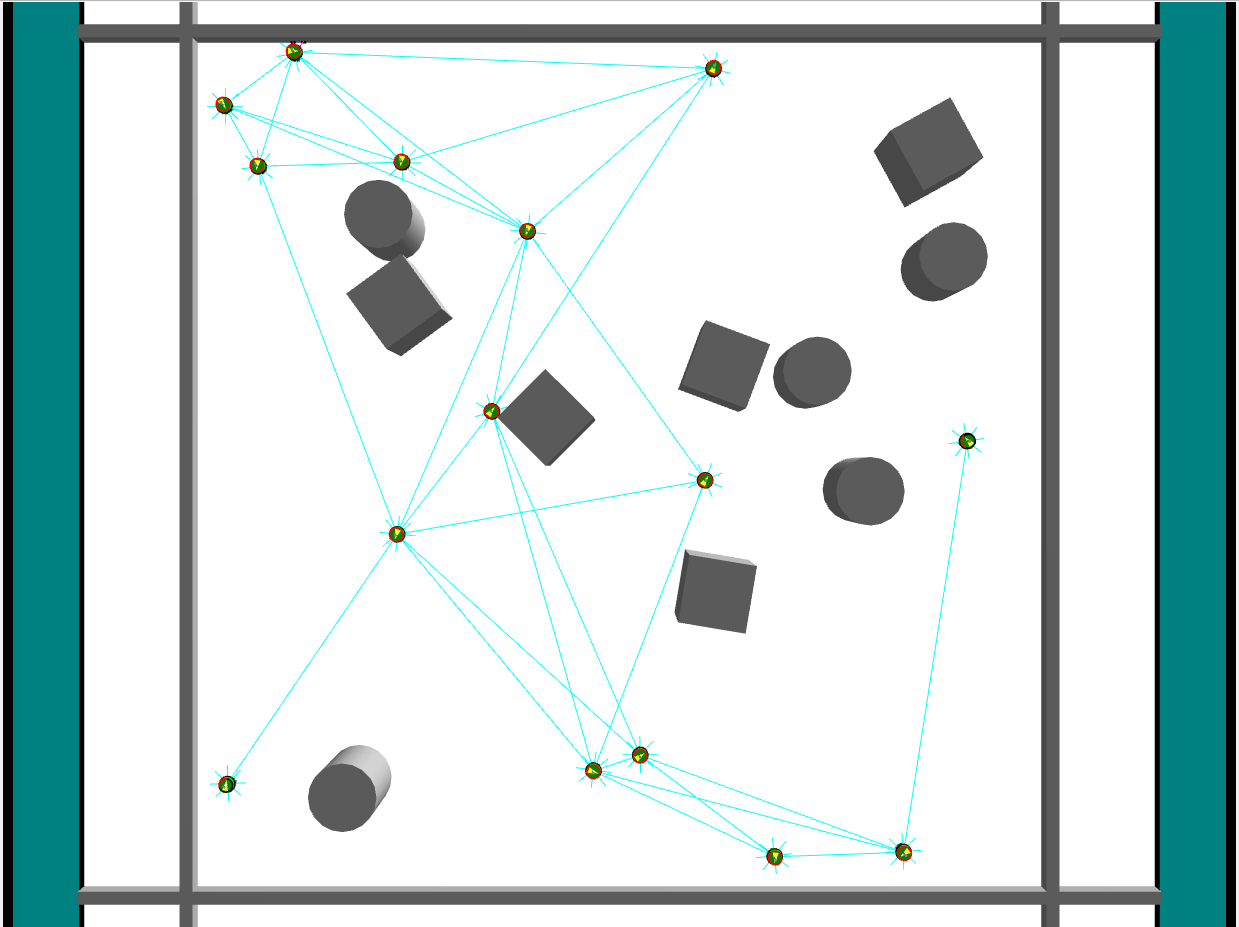}
    \vspace{-0.5cm}
    \caption{Experimental arena. The Pi-puck robots (small green circles) are connected by cyan lines when they are able to record each other's trajectories through their range-and-bearing sensors. The short cyan rays around each robot represent collision avoidance sensors used to detect walls (delimiting the arena), obstacles (gray cubes and cylinders), and other robots.}
    \label{fig:arenaconf}
 \end{SCfigure}

Our swarm of 15 robots navigates a $5\times5$\,m$^2$ arena, while avoiding obstacles and other robots. Obstacles---5 cylinders with $0.15$\,m radius and 5~boxes with $0.3\times0.3\,$m$^2$ base---are distributed uniformly at random in the arena. Each experiment has a duration of 5\,000\,s. A visual representation of the arena can be seen in Fig.~\ref{fig:arenaconf}.

\subsection{Robot Model and Controller}

The robots we use in the simulator are the Pi-pucks~\cite{MilJoyHil-etal2017:iros,MonBonRae-etal2009:icarsc}, which have been demonstrated in previous research to be capable of executing Ethereum block\-chain software~\cite{PacStrDor2020:ants,StrPacDor2023:sciencerobotics}.
The Pi-pucks are equipped with an array of infrared sensors to detect obstacles, a range-and-bearing board to obtain data regarding the motion and identities of the other robots, and a Wi-Fi module to synchronize the blockchain blocks and transactions. All communications occur within a maximum range of 2.5\,m.
 
Each robot moves in the environment, while avoiding other robots and obstacles, and collects data regarding its own motion and the motion of its peers using a range-and-bearing sensor. A motion trajectory is a list of $x,y$ positions over 10\,s. If the connection with another robot is interrupted during the recording, the corresponding  trajectory is discarded.

Every 100\,s, a robot enters a training state. In this state, it collects the last aggregated model from the blockchain. This model is then trained on the latest collected data using Tensorflow~\cite{AbaBarChe-etal2016:usenix}. The parameters used are the same as in~\cite{MajSriPin2021:icra}: a batch size of 20, a learning rate of 0.001 and 20~epochs.

Once the new local model is trained, it is included in a blockchain transaction and broadcast so that it may contribute towards the next iteration of the shared model. Immediately after, each robot begins collecting the next set of data.

\subsection{Model Aggregation}

The aggregation of the local models is performed using a smart contract whose pseudocode is given in Algorithm~1. It contains two functions that the robots can interact with:%
\begin{itemize}
    \item \texttt{submitModel}: robots can upload their locally trained models, which will be used for updating the shared model using the FedAvg algorithm~\cite{McMMooRam-etal:2017}. The shared model is updated each time a certain fraction, i.e., a \textit{quorum}, of the robots have submitted their models. We selected a quorum size of $\lfloor q \times N_r \rfloor$ robots, with $q=50\%$  and $N_r$ being the number of robots in the swarm. This results in a quorum of 7 out of 15 robots. If the quorum size is too large, the learning process can be slowed down by robots that fail to submit their models (e.g., if they are not connected)~\cite{McMMooRam-etal:2017}. Conversely, if it is too small, it may pose a security vulnerability as the Byzantine robots might become the majority in a single aggregation round.
    \item \texttt{getModel}: robots can retrieve the latest shared model and use it for local training. If a robot has not synchronized its blockchain with the latest version due to network partitioning, it may train its model using an outdated model. This issue can be addressed by selecting adequate parameters and making appropriate choices for the blockchain consensus protocol that accommodates network properties and robot capabilities. Given the speed and communication range of the robots, we used a block period of 10\,s. By waiting a minimum of 10\,s between the generation of blocks, we grant enough time for robots to synchronize the previous blocks and therefore the latest state of the smart contract.
\end{itemize}

The shared model is calculated using a weighted average---based on the number of samples that the robots used to train their model---of the local models submitted by the robots in that aggregation round's quorum. In this way, models that were trained using more data have a larger impact on the shared model. This adaptation of FedAvg~\cite{McMMooRam-etal:2017} is shown in Algorithm~1.

\begin{algorithm}[t]
\label{alg:sc}
    \caption{\textbf{\textit{FedAvg}} with $N_r=15$ total robots; $q=50\%$ quorum of robots participating; $B=20$ local mini-batch size; $E=20$ local epochs; $\mathcal{P}_k$ data of robot $k$, $R$ aggregation rounds, loss function $\ell$ (mean square error), and learning rate $\eta=0.001$.}
    \label{fedavgalgo}
    \textbf{Init:} model weights $w(0)$\\
    \textbf{Smart contract executes:}
    \begin{algorithmic}
        \For{round $t = 1$ to $R$}
            \State $p \leftarrow max(\lfloor q \cdot N_r\rfloor, 1)$
            \State $S_t \leftarrow$ (random set of $p$ robots)
            \State $N_s \leftarrow$ 0
            \ForAll{robot $k \in S_t$}
                \State $n_k(t+1) \leftarrow$ (number of samples of robot $k$)
                \State $N_s \leftarrow N_s + n_k(t+1)$
                \State $w_{k}(t+1) \leftarrow$ RobotTrain$(k, w(t))$
            \EndFor
            \State $w(t+1) \leftarrow  \frac{1}{N_{s}} \sum^{p}_{k=1}  n_k \cdot w_{k}(t+1)$
        \EndFor \[\]
    \end{algorithmic}
    \textbf{RobotTrain$(k, w)$:}
    \begin{algorithmic}
        \State $\mathcal{B} \leftarrow$ (split $\mathcal{P}_k$ in batches of size $B$)
        \For{epoch $e = 1$ to $E$}
            \For{batch $b \in \mathcal{B}$}
                \State $w \leftarrow w - \eta\nabla \ell(b;w)$
            \EndFor
        \EndFor 
    \Return $w$
    \end{algorithmic}
\end{algorithm}

\subsection{Security Mechanisms}
\label{sec:securitymechanism}

In the federated learning literature, security mechanisms are typically studied in a centralized setting (see Section~\ref{sec:background}), in which a model aggregation server is in charge of qualifying and rejecting the individual models, as well as of managing the identities of the participants and protecting from Sybil attacks. In the following, we describe how we implemented decentralized security mechanisms.

\paragraph{Sybil protection}

In order to protect the system against Sybil attacks (i.e., situations in which an attacker forges many identities in order to gain control over the swarm), robots must transfer 5~crypto tokens (i.e., scarce units that are maintained on the blockchain) in order to submit their models to the smart contract. Robots start with 21~tokens and if they remain with less than 5 tokens, they can no longer participate in federated learning. The tokens thus act as participation credentials which are allocated through a blockchain-based reputation system that evaluates the quality of the models submitted by the robots.

\paragraph{Outlier rejection}

Most security measures in the federated learning literature begin with rejecting outliers, as it is a simple yet effective method to prevent the shared model from diverging when individual learners submit extreme inputs. We employ a static threshold, which is compared to the distance between the submitted model weights $w_s$ and the model weights $w_a$ aggregated in the last round:%
\begin{equation}
    \frac{1}{N}\sum^{N}_{i=1} |w_{s,i} - w_{a,i}| \leq 0.05\,,
    \label{firstsecurityeq}
\end{equation}
The total number of trainable model weights is $N=2\,848$, in our case. In a pilot study, we observed that, on average, the distance between the submitted model and the shared model was $0.01$, so we established the threshold to be 5~times this value. If a robot submits a model with weights that surpass the threshold, the model will be excluded from the aggregation and the robot that submitted it loses the tokens paid for the submission.

\paragraph{Ranking system}
The ranking systems sorts the models submitted by the robots according to their distance from the shared model. The number of ranked models in each aggregation round corresponds to the quorum to be reached to trigger the model aggregation. In our case, the quorum is 7 robots and therefore we rank 7 models. The model with the distance that corresponds to the median of all the distances is ranked first. Then the following ranks are based on the absolute difference between a model's distance and the first-ranked model's distance. 

The tokens sent by the robots alongside their model submissions are added to a reward pool, that is redistributed once the shared model is updated. Robots that submitted a model with a higher rank receive a larger reward than those that submitted a model with a lower rank, according to a reward weights vector. The reward weights vector we use is $k = [1, 1, 1, 1, 1, -1, -1]$, which means that the robots that submitted the two worst-ranked models are penalized and lose tokens, whereas the others are rewarded. In addition, we scale the rewards proportionally to the number of data samples used to train the model, because using more samples normally improves the training outcome but also increases the training and data-collection costs. To perform this scaling, we first divide the robots into a group that is rewarded $\mathsf{R}$ and a group that is penalized $\mathsf{P}$ (the penalized group gets back fewer tokens than submitted). Then, each robot~$i$ receives the following quantity of tokens:
\[
R_i = 
\begin{cases} 
(1 + \frac{k_i \cdot s_i}{\sum\limits_{j\in\mathsf{R}}k_j \cdot s_j}) \cdot 5 \, \text{tokens} & \text{if } i \in \mathsf{R} \\
(1 - \frac{k_i \cdot s_i}{\sum\limits_{j\in\mathsf{P}}k_j \cdot s_j}) \cdot 5 \, \text{tokens} & \text{if } i \in \mathsf{P},
\end{cases}
\]%
where $k_i$ and $s_i$ are the reward weight and the number of samples submitted by robot $i$, respectively.
By using this reward function we can ensure that:%
\begin{enumerate}
    \item High rank and high number of samples lead to high reward.
    \item High rank and low number of samples lead to low reward.
    \item Low rank and low number of samples lead to low penalization.
    \item Low rank and high number of samples lead to high penalization.
\end{enumerate}

\subsection{Byzantine robots}
\label{sec:byzbehaviors}

We consider three types of Byzantine robots: faulty, malicious, and smart.
\paragraph{Faulty}

The first type of Byzantine robot is considered to be \textit{faulty}. It sends random model weights that are uniformly distributed in the interval between -0.5 and 0.5, which is the interval TensorFlow~\cite{AbaBarChe-etal2016:usenix} uses when generating random model weights at initialization. 
In a real-world deployment, such a behavior may occur if a robot's sensor is obstructed or damaged and thus yields arbitrary values.

\paragraph{Malicious}
The second type of Byzantine robot is considered to be a \textit{malicious} robot that sends as its trained model the model from the previous aggregation round, stored in the blockchain. Such a behavior could occur because the robot is hacked, and the hacker is attempting to bypass the outlier rejection threshold and slow down the training.

\paragraph{Smart}
Finally, the third type of Byzantine robot is a \textit{smart} robot. Its objective is to achieve high ranks by exploiting the ranking system in order to increase the number of participation tokens it holds. To do so, it predicts the most likely next update to the shared model and submits this model, without performing any model training or data collection. More specifically, a smart robot $z$ computes how much the weights changed on average during the last aggregation step (from $t-1$ to $t$) and applies a random variation of the same magnitude to its current model weights $w(t)$. The weights the smart robot sends are:
\begin{equation}
    w_{z,i}(t+1) = w_{i}(t) + 2r \cdot \Delta(w(t),w(t-1))\,,
\end{equation}
\begin{equation*}
\text{where } \Delta(w(t),w(t-1)) = \frac{1}{N}\sum^{N}_{i=1} |w_{i}(t) - w_{i}(t-1)|\,,
\end{equation*}%
and $r$ is a uniformly distributed random value between 0 and 1 which is regenerated for every weight $w_{z,i}(t+1), i \in \{1,2,\ldots,2848\}$ in the robot's weights vector~$w_z$.

\subsection{Metrics}

\paragraph{Average Loss} This is the loss obtained by taking a weighted average of the validation loss of each robot's model taking part in the aggregation round.
Note that for all plots showing the average loss we use logarithmic scale. The reason for this is that in many cases our results for different configurations are very similar and would not be distinguishable on a linear scale.

\paragraph{Tokens gained} This is the number of tokens that robots have gained at the end of an experiment due to the rewards system. Depending on the experiment, we count separately the number of tokens gained by Byzantine robots and those gained by non-Byzantine robots. 

\section{Results}
\label{sec:results}

\subsection{Experiment 1 -- No security}
\label{exp1results}

Storing large amounts of data on hardware-limited robots is costly, so it is important to allow for old data to expire and instead use more recent data, as long as model quality is not compromised.
In this experiment, we initially employ 15~non-Byzantine robots and vary the \textit{data expiration time} from 250\,s to 1\,250\,s, to determine how it impacts the convergence speed of the model. This experiment serves as a baseline in order to compare the results to later experiments and to the results of Flow-FL. 

\begin{figure}[t]
    \centering
    \begin{subfigure}[t]{0.48\textwidth}
    \vskip 0pt
    \centering
        \includegraphics[width=\textwidth]{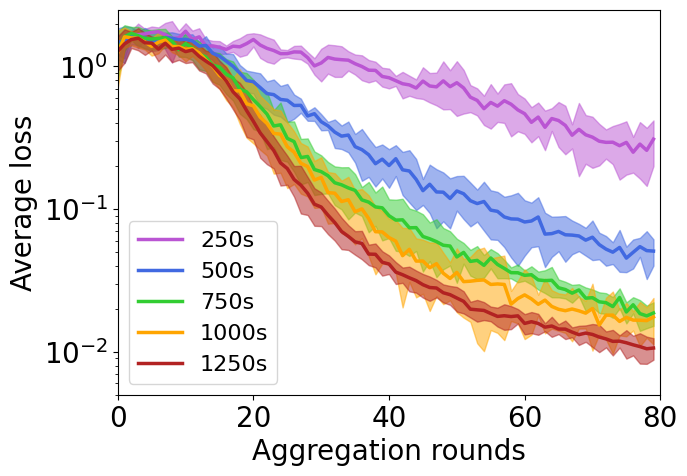}
        \caption{The effect of data expiration time on the average loss with no Byzantine robots.}
        \label{fig:exp1convspeed}
    \end{subfigure}\hspace*{1em}
    \begin{subfigure}[t]{0.48\textwidth}
    \vskip 0pt
    \centering
        \includegraphics[width=\textwidth]{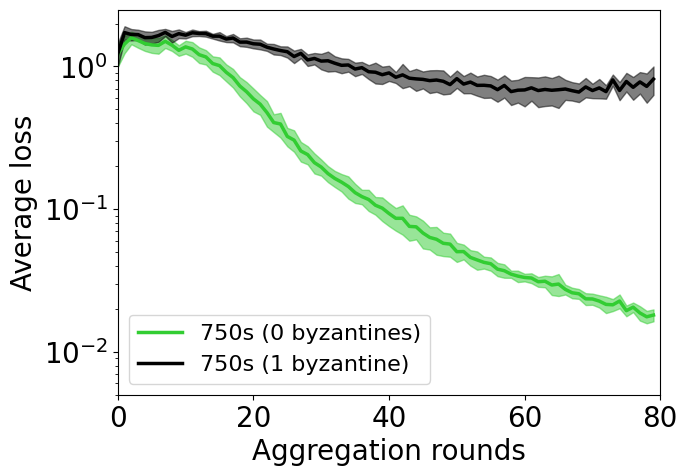}
        \caption{The effect of introducing one Byzantine robot without security mechanisms.}
        \label{fig:lossnosecurity}
    \end{subfigure}
    \caption{Each line shows the average loss of the aggregated model---5~runs for (a) and 10~runs for (b)---and the shaded areas depict 95\,\% confidence intervals. \textbf{(a)}~Increasing the data expiration time leads to a quicker decrease of the average loss. This is expected, as more data is used for model learning. However, it also shows that for data expiration times of 750\,s and higher, the loss curves are similar. For this reason, we select a value of 750\,s for all subsequent experiments.
    \textbf{(b)}~A single Byzantine robot is sufficient to prevent the model from converging (with a data expiration time of 750\,s).}
\end{figure}

Figure~\ref{fig:exp1convspeed} shows that increasing the data-expiration time beyond 750\,s only results in marginal improvement.
Therefore, we selected the 750\,s expiration time as the data expiration time for the subsequent experiments. Comparing our results with the ones obtained in the original article that proposed the Flow-FL framework \cite{MajSriPin2021:icra}, the convergence speed of our system (Figure \ref{fig:exp1convspeed}) is quantitatively different but qualitatively very similar: we reach the same final results with a loss of $10^{-2}$ for the higher expiration times, but with a slower convergence speed. This discrepancy may be caused by differences in the valuess of the parameters (as it was not possible to retrieve all the values used in the original Flow-FL article \cite{MajSriPin2021:icra}).

After selecting an appropriate data expiration time and showing that the system functions similarly to Flow-FL (but using a blockchain), we test the system's resilience to the introduction of a single \emph{faulty} Byzantine robot, without yet applying the security measures described in Section~\ref{sec:securitymechanism}. Figure~\ref{fig:lossnosecurity} shows that the system is vulnerable to even a single faulty Byzantine robot. For this reason, introducing security mechanisms is an obvious necessity.

\subsection{Experiment 2 -- With security}
\label{exp3results}

In this set of experiments, we use the smart contract that implements the security mechanisms explained in Section~\ref{sec:securitymechanism}. We conduct 20~runs for each configuration, ranging from 0 to 7 Byzantine robots.

\subsubsection{Security to faulty robots}

Initially, we introduce faulty Byzantine robots (see Section~\ref{sec:byzbehaviors}). We expect that a vast majority of incorrect models from faulty Byzantine robots will be rejected by the outlier rejection mechanism (see Section~\ref{sec:securitymechanism}). Therefore, these experiments mainly showcase the efficacy of this mechanism in handling faulty Byzantine behaviors, which are not malicious in nature.

\begin{figure}[t]
    \centering
    \begin{subfigure}[t]{0.48\textwidth}
        \vskip 0pt
        \centering
         \includegraphics[width=\textwidth]{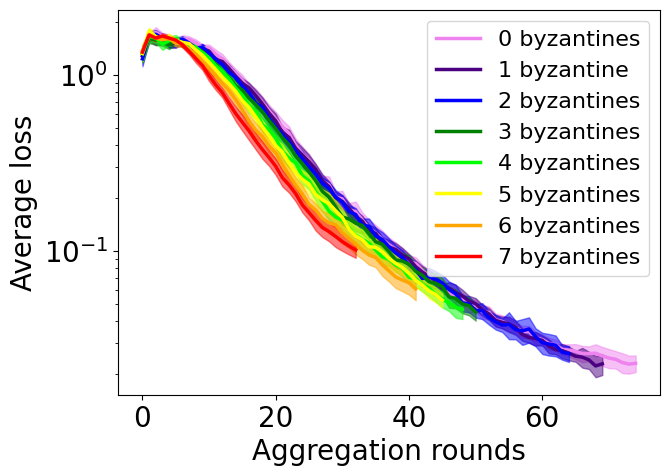}
        \caption{Average loss for different numbers of faulty Byzantine robots.}
        \label{fig:exp3loss}   
    \end{subfigure}\hspace*{1em}
    \begin{subfigure}[t]{0.48\textwidth}
        \vskip 0pt
        \centering
        \vspace{3mm}
        \includegraphics[width=\textwidth]{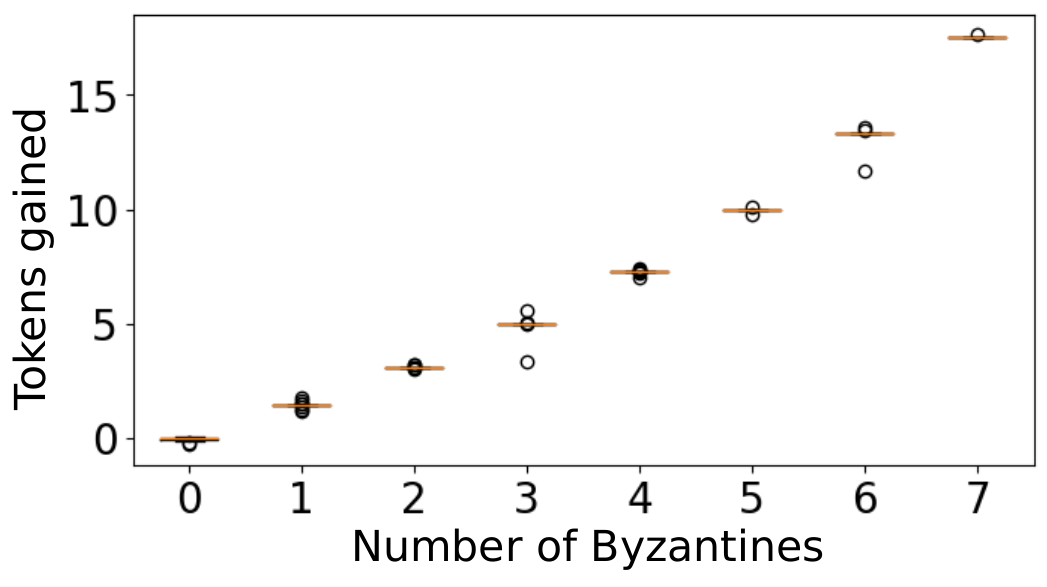}
        \vspace{2mm}
        \caption{Tokens gained by non-Byzantine robots.}
        \label{exp3averagegoodether}
    \end{subfigure}
        \caption{\textbf{(a)}~As the number of Byzantine robots increases, the number of aggregations that occur within the fixed 5\,000\,s experiment duration decreases. Indeed, by rejecting outlier models, we can achieve a more robust and faster convergence (in terms of aggregation rounds). However, fewer aggregation rounds occur by the end of the experiment as there are fewer robots producing reliable models and the quorum size is the same (7~robots). \textbf{(b)}~As the number of Byzantine robots increases, the non-Byzantine robots gain more tokens.}
        \label{fig:exp3goodether}
\end{figure}

Figure~\ref{fig:exp3loss} shows that the implemented security mechanisms are effective at rejecting inputs from faulty Byzantine robots and the resulting average loss is largely independent of the number of Byzantine robots. Since the secured smart contract requires 5~tokens for robots to submit their models, and since these tokens are later distributed among the robots whose models got accepted, we expect that non-Byzantine robots gain tokens, whereas Byzantine robots lose tokens. This is shown in Figure~\ref{exp3averagegoodether}: non-Byzantine robots indeed gain tokens on average, indicating that our security mechanism is effective in distributing the tokens supplied by the Byzantine robots among the non-Byzantine robots.

\subsubsection{Security to malicious robots}
\label{exp4results}

\begin{figure}[ht]
    \centering
    \includegraphics[width=.5\textwidth]{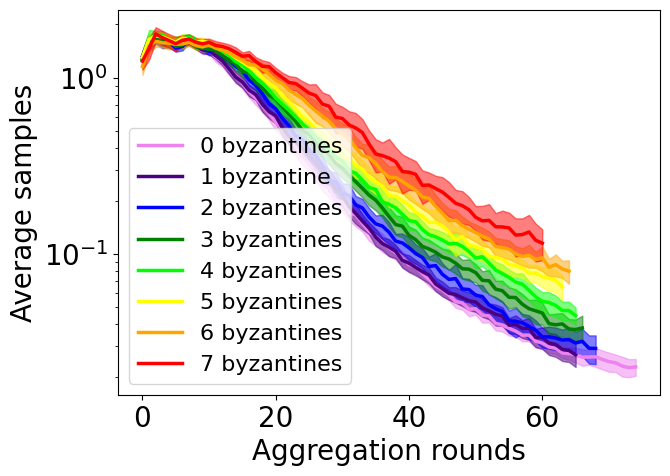}
    \caption{Average loss for different numbers of malicious Byzantine robots. Since a large number of models from malicious Byzantine robots are included in the aggregation, we do not observe the steep reduction in the total number of aggregation rounds as before. Even though the ranking mechanism secures the models' convergence, the model convergence becomes slower when the number of Byzantine robots increases.}
    \label{fig:exp4losscurve}
\end{figure}
\begin{figure}[ht]
    \centering
     \begin{subfigure}[t]{0.49\textwidth}
     \vskip 0pt
         \centering
         \includegraphics[width=\textwidth]{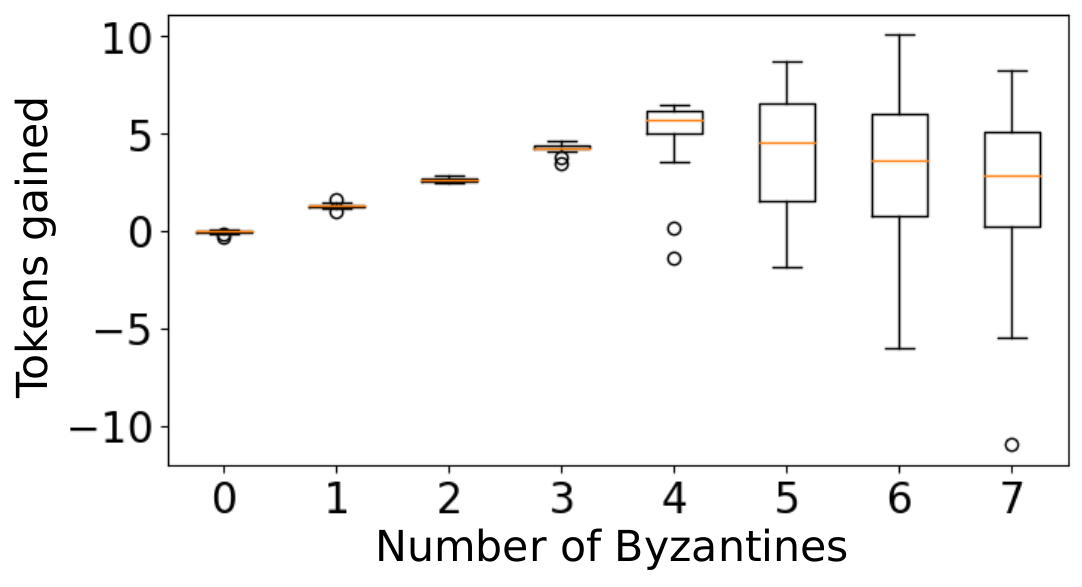}
         \caption{Tokens gained by non-Byzantine robots}
        \label{exp4averagegoodether}
     \end{subfigure}\hspace*{1em}
 \begin{subfigure}[t]{0.49\textwidth}
 \vskip 0pt
         \centering
         \includegraphics[width=\textwidth]{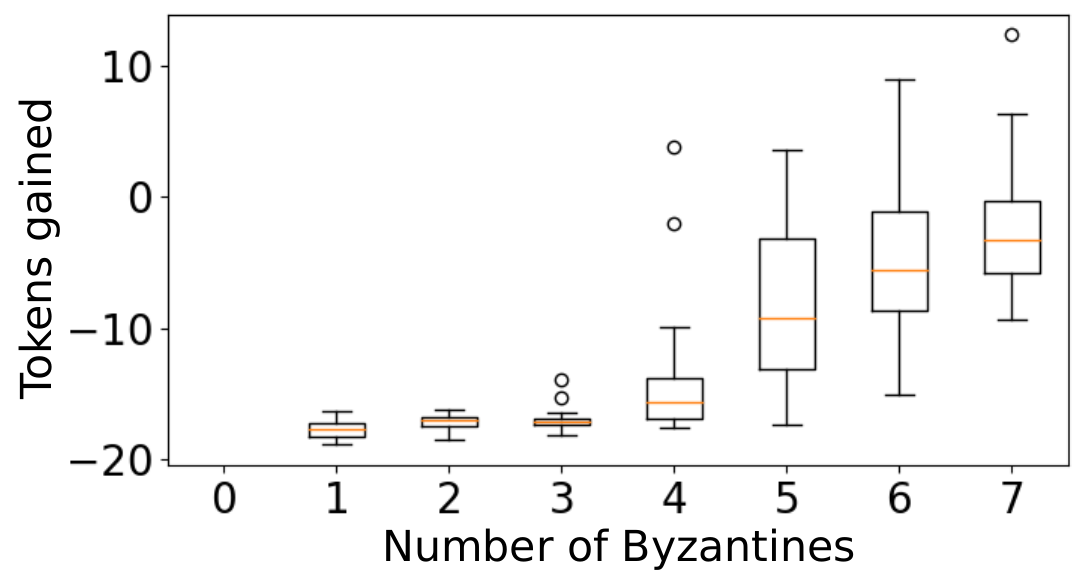}
         \caption{Tokens gained by Byzantine robots}
        \label{exp4averagebadether}
     \end{subfigure}     
    \caption{With up to 3 malicious Byzantine robots, our ranking mechanism successfully rewards non-Byzantine robots and penalizes Byzantine robots. With 4 Byzantine robots, this is no longer guaranteed, and with 5 or more of them, Byzantine robots will likely gain tokens.}
    \label{fig:exp4goodether}
\end{figure}

Although outlier rejection can protect the learning process from random outlier values, a malicious Byzantine robot could still attempt to manipulate the learning process by sending slightly deviated models that are yet within the established threshold. The impact of such an attack on the shared model is potentially smaller, since the model deviation is smaller; however, a collusion of malicious robots could succeed at manipulating the shared model. Our mechanism to manage this type of malicious Byzantine attack ranks the robot submissions according to how far they deviate from the median value of all submissions, and rewards the robots according to their rank (see Section~\ref{sec:securitymechanism}).

The results in Figure~\ref{fig:exp4losscurve} show that the implementation of the ranking system allows for model convergence, even in the presence of malicious Byzantine robots. However, when the number of Byzantine robots increases, the model convergence becomes slower and the loss at the end of the experiment is higher. Unlike faulty Byzantine robots, whose models were always rejected, the models from malicious robots are most of the time accepted, which leads to a slower convergence of the shared model, but also to faster aggregation rounds (note that in our experiments with a fixed duration of 5\,000\,s, the aggregation rounds when malicious robots were present are more than in experiments with faulty robots, compare Figures \ref{fig:exp3loss} and \ref{fig:exp4losscurve}). Figures~\ref{exp4averagegoodether} and~\ref{exp4averagebadether} show that the non-Byzantine robots gain tokens up to a maximum of 3 Byzantines. After this point, the ranking system starts to fail as the Byzantine robots can occasionally become a majority in a single aggregation round quorum (whose total size is 7 robots), which leads to Byzantine robots maintaining or even gaining tokens.

\subsection{Experiment 3 - Vulnerability to smart Byzantine robots}
\label{sec:experiment3}

By using crypto tokens as participation credentials we can protect the system from Sybil attacks and mitigate the negative impact of Byzantine robots. However, in a real-world operation, it is possible that robots temporarily exhibit faulty behavior, becoming Byzantine robots, and then recover. For this reason, we designed the token rewards system to not only penalize Byzantine behaviors, but also to reward robots that contribute with valid models. This ensures that the reputation tokens recirculate between the robots and do not run out, which would otherwise halt the system.
However, this introduces a common vulnerability to reputation-based systems~\cite{sabater2005reputation}: a smart Byzantine robot could attempt to obtain tokens by initially sending good models, and later use these tokens to manipulate the shared model.

We show this possibility by performing experiments using smart Byzantine robots that try to be ranked first by sending a model that follows the current trend of the training (see Section~\ref{sec:byzbehaviors}).\footnote{Simulations in this section are repeated 18~times due to time constraints.}

\begin{figure}[t]
    \centering
    \includegraphics[width=.5\textwidth]{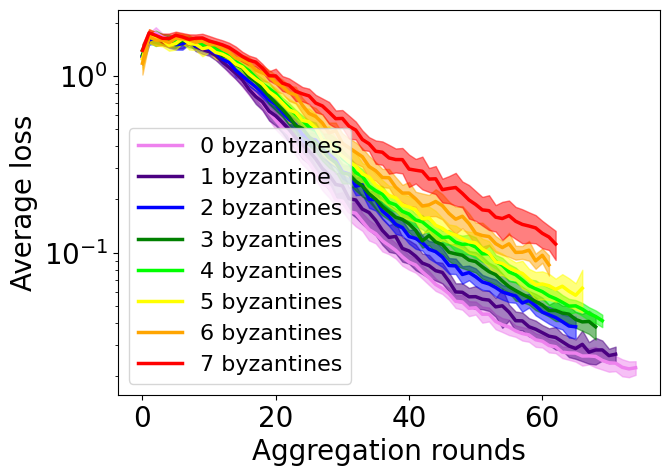}
    \caption{Average loss for different numbers of smart Byzantine robots. The model convergence speed and number of aggregation rounds does not differ much from the previous experiment shown in Figure \ref{fig:exp4losscurve}.}
    \label{fig:exp5losscurve}
\end{figure}

\begin{figure}[t]
     \begin{subfigure}[t]{0.49\textwidth}
         \vskip 0pt
         \centering
         \includegraphics[width=\textwidth]{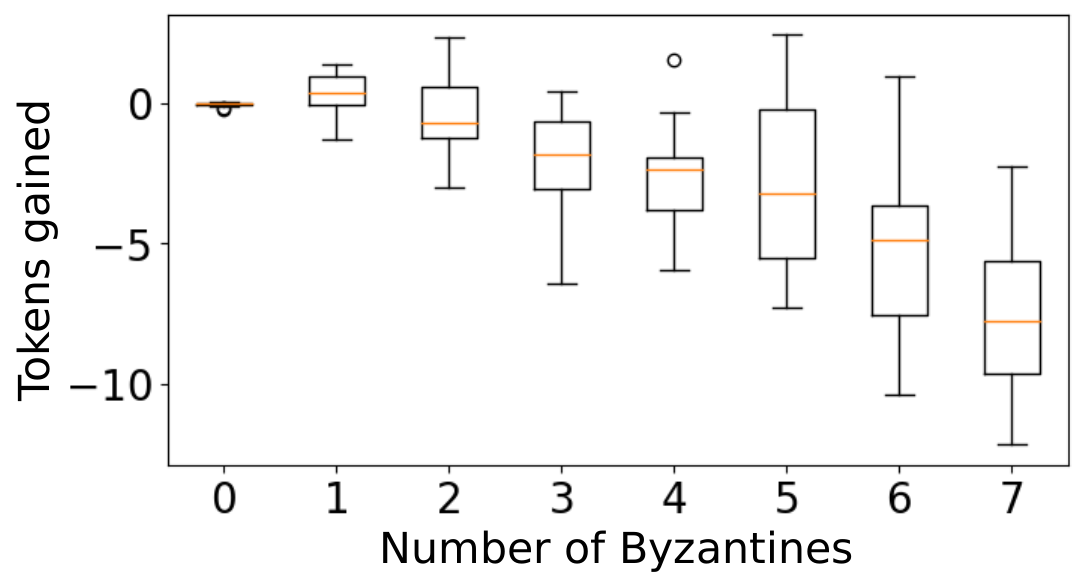}
         \caption{Tokens gained by non-Byzantine robots}
        \label{exp5averagegoodether}
     \end{subfigure}
~
          \begin{subfigure}[t]{0.49\textwidth}
              \vskip 0pt
         \centering
         \includegraphics[width=\textwidth]{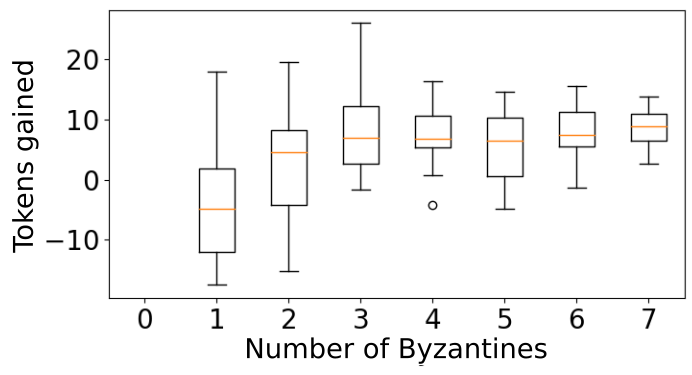}
         \caption{Tokens gained by Byzantine robots}
        \label{exp5averagebadether}
     \end{subfigure}
    \caption{Smart Byzantine robots are able to consistently gain more tokens than the non-Byzantine group, especially when there are 3 or more Byzantine robots.}
    \label{fig:exp5goodether}
\end{figure}%
The loss curve in Figure~\ref{fig:exp5losscurve} follows the same pattern as the loss curve observed in Figure~\ref{fig:exp4losscurve}: with a higher number of smart Byzantines, the training becomes slower. Still, with up to 4 smart Byzantine robots, the number of aggregation rounds does not decrease significantly. However, with a higher number of smart Byzantine robots, non-Byzantine robots may begin to lose reputation tokens thus reducing the number of learners and, therefore, slowing down the aggregation process, which then falls under the control of the smart Byzantine robots. This is more clearly visible in Figure~\ref{fig:exp5goodether} which shows how the smart Byzantine robots are able to gain more tokens than the non-Byzantine robots. Although the effect on model convergence is not visible (Figure~\ref{fig:exp5losscurve}), the smart Byzantine robots, as soon as they detain a majority of the tokens, can potentially switch behavior and collude on manipulating the shared model.

\section{Discussion and Conclusions}
\label{sec:discussion}

Federated learning is a promising approach to improve machine learning in distributed systems. However, its application in the context of swarm robotics is hampered by the need for infrastructure that establishes secure aggregation of the individual machine learning models. We showed that a possible solution is to incorporate blockchain technology into robot swarms to establish a secure federated learning framework that respects the decentralization requirement and employs security mechanisms that are a must for real-world deployments.

Our experiments showed that Byzantine robots pose a significant threat to federated learning, hindering or potentially manipulating the training of the shared model. We thus implemented security mechanisms to mitigate this threat. Our results show that these mechanisms worked as intended and enabled the learning process to maintain good results despite the presence of Byzantine robots.
The proposed mechanisms can also effectively protect the system from Sybil attacks by limiting the number of times a robot is allowed to submit model weights that are labeled as outliers. As faults naturally occur in robots over time, robots would be at the risk of eventually exhausting all participation tokens.
To allow the system to autonomously recover from temporary faults (i.e., without the need for a technician and administrator to diagnose the robots and grant them new participation tokens), we established a reputation system that not only penalizes, but also rewards robots when they submit good-quality models.
However, our results also show how this scheme is vulnerable to smart Byzantine robots that attempt to gain tokens without performing any useful work. Future work needs to address this issue.

The costs of implementing our method should also be discussed in terms of storage and bandwidth usage. Aggregating the models in the blockchain requires that robots broadcast a large number of transactions that include the trained models, which can be several KB in size. Regarding storage costs, since each blockchain node stores the transactions in blocks, every aggregated weight submitted by a robot will be stored on the blockchain. Experimentally, we observed that the storage cost increases linearly with time (the blockchain size increases by approximately 16.7 KB each time a robot submits its model through a transaction) and reaches on average approximately 100~MB at the end of an experiment (i.e., after 5,000 seconds) training an artificial neural network with 2848 weights. In this sense, blockchains can be seen as storage-efficient data structures that scale well, given the storage capabilities of the Pi-Pucks (16~GB with a default SD card). Still, in swarm robotics, their costs might impede their use on less capable robots.

Our work is one of the very first steps~\cite{MajSriPin2021:icra,XIANJIA2021135,10025836,10105168} towards the successful deployment of decentralized federated learning in swarm robotics, enabling robot swarms to agree on global information without compromising their critical properties: autonomy, decentralization, and scalability. Although this proof of concept has demonstrated the potential of the proposed solution, further research is needed to refine and enhance it, for example, in terms of security to more sophisticated attacks.

\subsubsection*{Acknowledgements.}
The authors thank Carlo Pinciroli and Nathalie Majcherczyk for sharing the data and the code of their original Flow-FL paper. M.D. and V.S. acknowledge support from the Belgian F.R.S.-FNRS. A.R. acknowledges support from DFG under Germany's Excellence Strategy - EXC 2117 - 422037984.

\bibliographystyle{splncs04}
\bibliography{mybibliography,mybibliographyvolker}

\end{document}